\documentclass[twocolumn]{el-author}

\newcommand{\ua}{\uparrow}
\newcommand{\nc}{\newcommand}
\nc{\da}{\downarrow} \nc{\hc}{\hat{c}} \nc{\hS}{\hat{S}}
\nc{\bra}{\langle} \nc{\ket}{\rangle} \nc{\eq}{equation (\ref}
\nc{\h}{\hat} \nc{\hT}{\h{T}}\nc{\be}{\begin{eqnarray}}
\nc{\ee}{\end{eqnarray}}\nc{\rd}{\textrm{d}}\nc{\e}{eqnarray}\nc{\hR}{\hat{R}}\nc{\Tr}{\mathrm{Tr}}
\nc{\tS}{\tilde{S}}\nc{\tr}{\mathrm{tr}}\nc{\8}{\infty}\nc{\lgs}{\bra\ua,\phi|}\nc{\rgs}{|\ua,\phi\ket}
\nc{\hU}{\hat{U}}\nc{\lfs}{\bra\phi|}\nc{\rfs}{|\phi\ket}\nc{\hZ}{\hat{Z}}\nc{\hd}{\hat{d}}\nc{\mD}{\mathcal{D}}
\nc{\bd}{\bar{d}}\nc{\bc}{\bar{c}}\nc{\mc}{\mathcal}\nc{\ea}{eqnarray}\nc{\mG}{\mathcal{G}}\nc{\bce}{\begin{center}}
\nc{\ece}{\end{center}}

\begin{document}

\title{Fast LLMMSE filter for low-dose CT imaging}

\author{Fengling Wang, Bowen Lin, Shujun Fu, Shiling Xie, Zhigang Zhao, Yuliang Li}

\abstract{Low-dose X-ray CT technology is one of important directions of current research and development of medical imaging equipment. A fast algorithm of blockwise sinogram filtering is presented for realtime low-dose CT imaging. A nonstationary Gaussian noise model of low-dose sinogram data is proposed in the low-mA (tube current) CT protocol. Then, according to the linear minimum mean square error principle, an adaptive blockwise algorithm is built to filter contaminated sinogram data caused by photon starvation. A moving sum technique is used to speed the algorithm into a linear time one, regardless of the block size and the data range. The proposed fast filtering gives a better performance in noise reduction and detail preservation in the reconstructed images, which is verified in experiments on simulated and real data compared with some related filtering methods.}

\maketitle

\section{Introduction}

As an important medical imaging mode, X-ray computed tomography (CT) is one of common clinical diagnosis methods \cite{hsieh2009computed}. With wider applications of X-ray CT scan in examining human body, its radiation hazard to public health has received more and more attention \cite{brenner2007computed}. In face of increasingly severe radiation threat, the International Organization for Medical Physics (IOMP) has made quality control standards in medical radiation, advocating that the X-ray diagnosis should follow the principle of legitimacy in practice and protection optimization (As Low As Reasonably Achievable, ALARA) \cite{slovis2003children}. Aiming to obtain best diagnostic effect at a minimum cost and radiation dose, low-dose CT technology has become one of main directions of current research and development of medical imaging equipment.

Low-dose CT scanning poses a huge challenge for CT reconstruction method. Under the low-dose condition the photon starvation appears at detector bins of CT system \cite{hsieh2009computed}, which leads to heavily noisy projection data and consequent degraded CT images full of heavy streak artifacts and noise when reconstructed from traditional algorithms. Thus, researchers have made great efforts from two aspects of low-dose CT reconstruction and image postprocessing. In the respect of low-dose CT reconstruction, there are three strategies for robust image reconstruction.  A strategy is the filtered backprojection (FBP) \cite{hsieh2009computed} based on filtering of noisy projection data \cite{wang2006noise,zhang2010statistical,wang2015adaptive}. In this letter, a fast blockwise algorithm with the local linear minimum mean square error (LLMMSE) estimation is built to filter noisy sinogram data, which will be described in detail below. Another strategy is the statistics-based iterative image reconstruction (SIIR) algorithm \cite{wang2009iterative}. The third strategy is the few-view sparsity reconstruction (FVSR) algorithm based on sparse angle sampling \cite{sidky2006accurate,yu2009compressed}. As for low-dose CT image postprocessing, the subsequent processing based on property analysis and filtering of noise and artifacts in the reconstructed image is also an important direction of low-dose CT imaging \cite{ma2011low,chen2013improving}. By enhancing effectively the signal-to-noise ratio of collected data and the density resolution of CT images, these algorithms are expected to obtain CT images with quality equal to or higher than those of scanning with regular dose under the condition of lower dose.

\section{Noise modeling for sinogram data}

In the field of CT imaging, the calibrated and log-transformed projection data are called as the sinogram. For noise reduction algorithms in sinogram domain, noise modeling is of great importance for improving their performance, especially for low-dose CT imaging. Based on the previous study \cite{li2004nonlinear}, we propose the following noise model for low-dose sinogram data
\begin{equation}\label{equ:noisemodel}
    q=p+n=p+\sqrt{f\times\exp(p/\eta)}u,
\end{equation}
where $p$ and $q$ denote original noise-free and noisy sinogram data, respectively; $u$ denotes the Gaussian noise with zero mean and unitary standard deviation; $f$ and $\eta$ are object-independent parameters that are specified by different CT systems. $f$ is an adjustable factor adaptive to each detector bin. Obviously, the original sinogram data are contaminated by a nonstationary Gaussian noise $n$ ($=\sqrt{f\times\exp(p/\eta)}u$) caused by the photon starvation under the low-dose condition.

\section{Blockwise filtering with LLMMSE estimation}

Considering the sinogram data model (\ref{equ:noisemodel}), if we impose a linear constraint on the estimation of the original data, we have the linear minimum mean square error (LMMSE) estimator \cite{kuan1985adaptive}
\begin{equation}\label{equ:lmmse}
    \widehat{p}_{LMMSE}=E(p)+C_{pq}C^{-1}_q(q-E(q)),
\end{equation}
where $E(p)$ and $E(q)$ are the ensemble means of $p$ and $q$, respectively; $C_{pq}$ is the cross-covariance of $p$ and $q$; $C^{-1}_q$ is the inversion of the covariance of $q$.

In order to compute nonstationary ensemble statistics, we propose a blockwise filtering of noisy sinogram data, and refer to it as a blockwise local linear minimum mean square error (LLMMSE-B) filter. Assuming uncorrelated noise and the nonstationary mean and nonstationary variance (NMNV) image model \cite{kuan1985adaptive}, for a point $(i,j)$ belonging to a block $w_k$ centered at the point $k$, the LLMMSE filter is defined as
\begin{equation}\label{equ:llmmse}
    \widehat{p}_{LLMMSE}(i,j)=\overline{q}(i,j)+\frac{v_q(i,j)-\sigma^2_n(i,j)}{v_q(i,j)}(q(i,j)-\overline{q}(i,j)),
\end{equation}
where $\overline{q}$ and $v_q$ are the local spatial mean and variance of $q$; $\sigma^2_n$ is local spatial variance of nonstationary noise $n$.

However, a point $(i,j)$ is involved in all the overlapping blocks $w_k$ that covers $(i,j)$. For different values of $\widehat{p}_{LLMMSE}(i,j)$ computed in different windows, we average all the possible values of $\widehat{p}_{LLMMSE}(i,j)$. Let $a(i,j)=\frac{v_q(i,j)-\sigma^2_n(i,j)}{v_q(i,j)}$, and $b(i,j)=(1-a(i,j))\overline{q}(i,j)$, (\ref{equ:llmmse}) becomes
\begin{equation}\label{equ:llmmse1}
    \widehat{p}_{LLMMSE}(i,j)=a(i,j)q(i,j)+b(i,j).
\end{equation}
Thus, the blockwise filtering with LLMMSE estimation is computed by
\begin{equation}\label{equ:llmmse2}
    \widehat{p}_{LLMMSE-B}(i,j)=\frac{1}{|w|}\sum_{k|(s,t)\in w_k}(a(s,t)q(i,j)+b(s,t)),
\end{equation}
where $|w|$ is the number of points in $w_k$.
Duo to the symmetry of blockwise computing, (\ref{equ:llmmse2}) can be rewritten as
\begin{equation}\label{equ:llmmseb}
    \widehat{p}_{LLMMSE-B}(i,j)=\overline{a}(i,j)q(i,j)+\overline{b}(i,j),
\end{equation}
where $\overline{a}$ and $\overline{b}$ are the local spatial means of $a$ and $b$, respectively.

In order to efficiently compute local spatial means in the proposed method, the box filter is exploited using a moving sum technique, which speeds the algorithm into a linear time one, regardless of the block size and the data range.

Based on what has been discussed above, we summarize our proposed algorithm as follows. The noisy sinogram data $q$ is filtered by LLMMSE-B to obtain a better estimate $\widehat{p}_{LLMMSE-B}$. Then, a low-dose CT image is reconstructed from filtered sinogram data $\widehat{p}_{LLMMSE-B}$ with the classic FBP method.

\section{Experiments and analyses}
First, we perform computer simulations to verify our method. The simulated sinogram data are produced by projecting a 2-D modified Shepp-Logan head phantom ($256\times256$) using the fan-beam ray-driven algorithm \cite{fesslerirt}. As described in \cite{zhang2010statistical}, the size of sinogram is $888\times984$, where $888$ and $984$ are numbers of detector bin and angle sample, respectively. The noisy sinogram data for low-dose CT are simulated by adding nonstationary Gaussian noise to noise-free sinogram, where the variance of the nonstationary Gaussian noise is determined by the exponential relationship following the formula (\ref{equ:noisemodel}). In this study we take $f=22500$ and $\eta=22000$ to simulate a case of lower tube current (mAs).

In the proposed filtering, the parameters are chosen as follows: $3\times3$ mask for calculating both median and mean values; local noise variance is 0.8 times the noise estimated according to the formula (\ref{equ:noisemodel}) using the means filter; a Hanning filter with default settings is employed in the fan-beam FBP reconstruction \cite{fesslerirt}.

\begin{figure}[ht]
 \centering
 \includegraphics[width=0.45\textwidth]{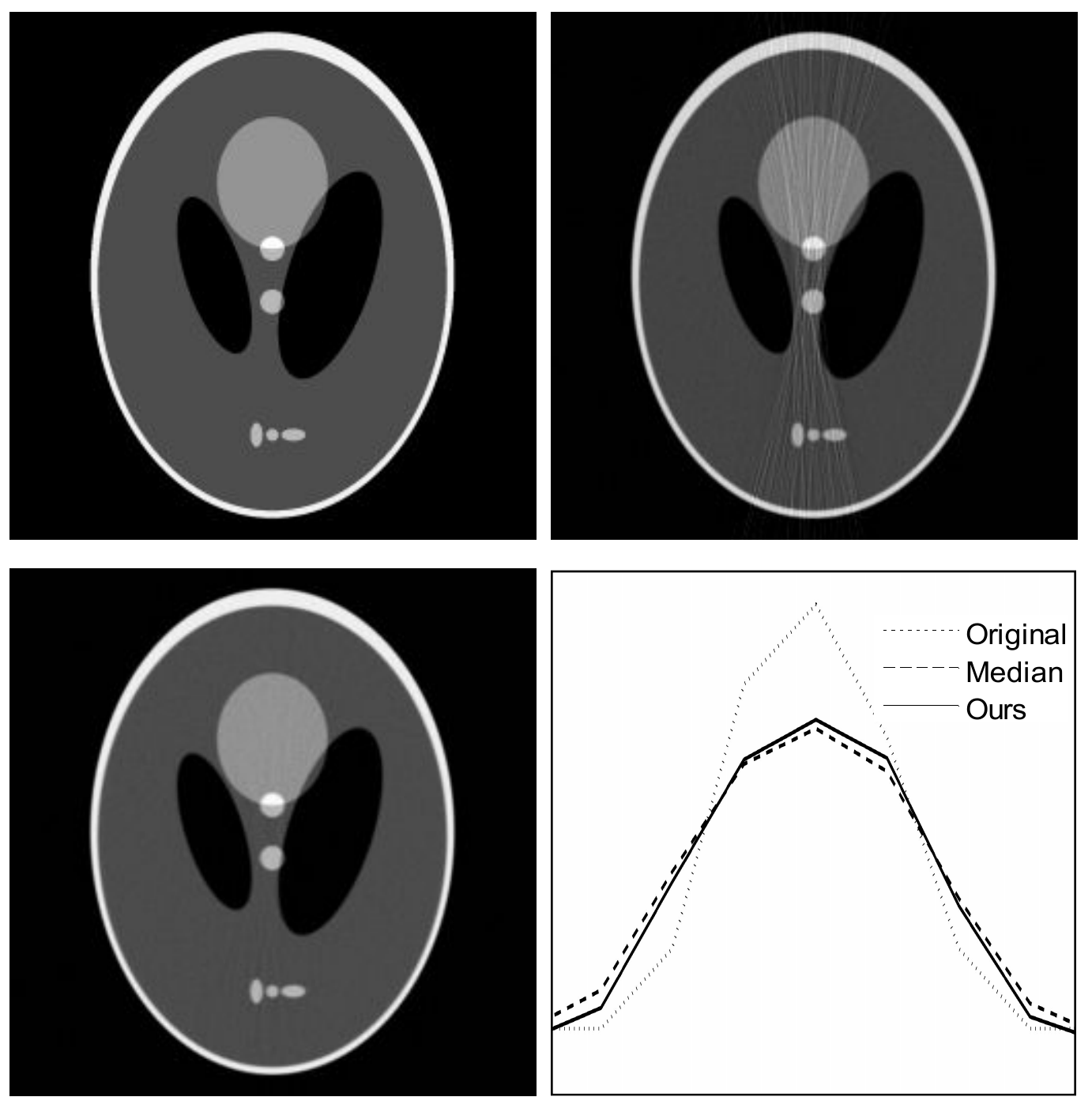}
 \caption{FBP reconstructed images from filtered sinogram data (from top-left to bottom-right): original head phantom, results from noisy and filtered sinogram data by our method, and local comparison of profiles (203-209 rows, 126th column) in reconstructed images from filtered sinograms by both median and our processings, respectively.}\label{fig:phantom1}
\end{figure}

As shown in Fig.\ref{fig:phantom1}, the FBP reconstructed image from noisy sinogram data is obviously full of noise and streak artifacts, compared with the image from the filtered sinogram by our method, which produces fewer annoying artifacts. Furthermore, observing local profiles (203-209 rows, 126th column) in reconstructed images from filtered sinograms by both the median and our processings, one can see that our result displays a better approximation to the original one and sharper edges with smaller edge width than the result by the median filter.

To further validate our method, we calculate the signal to noise ratios ($SNR$) of reconstructed images. We also record the running time of our proposed filtering, where the time of FBP reconstruction is not included. All methods are implemented using the MATLAB programming on a desktop computer with Intel Core(TM)2 Quad 2.83 GHz CPU and 4.00 GB Memory. In Table 1, we compare the reconstructed images using such filtering methods as the median filter ($MED$), the LLMMSE estimator ($LLMMSE$) and our method. One can see that, our proposed method has highest $SNR$ with a little bigger time consumption.

\begin{table}[!htbp]
 \processtable{$SNR$ results and $Running\ time$ of sinogram filtering for low-dose FBP reconstruction by related methods.}
 {\begin{tabular}{|l|c|c|c|c|}\hline
 &Noisy&$MED$&$LLMMSE$&Ours\\\hline
 $SNR$&23.5726&24.3442&24.6771&24.8756\\\hline
 $Time$(sec.)&-&0.0783&0.1045&0.1219\\\hline
 \end{tabular}}{}\label{tab:snrtime}
\end{table}

\begin{figure}[!htbp]
 \centering
 \includegraphics[width=0.47\textwidth]{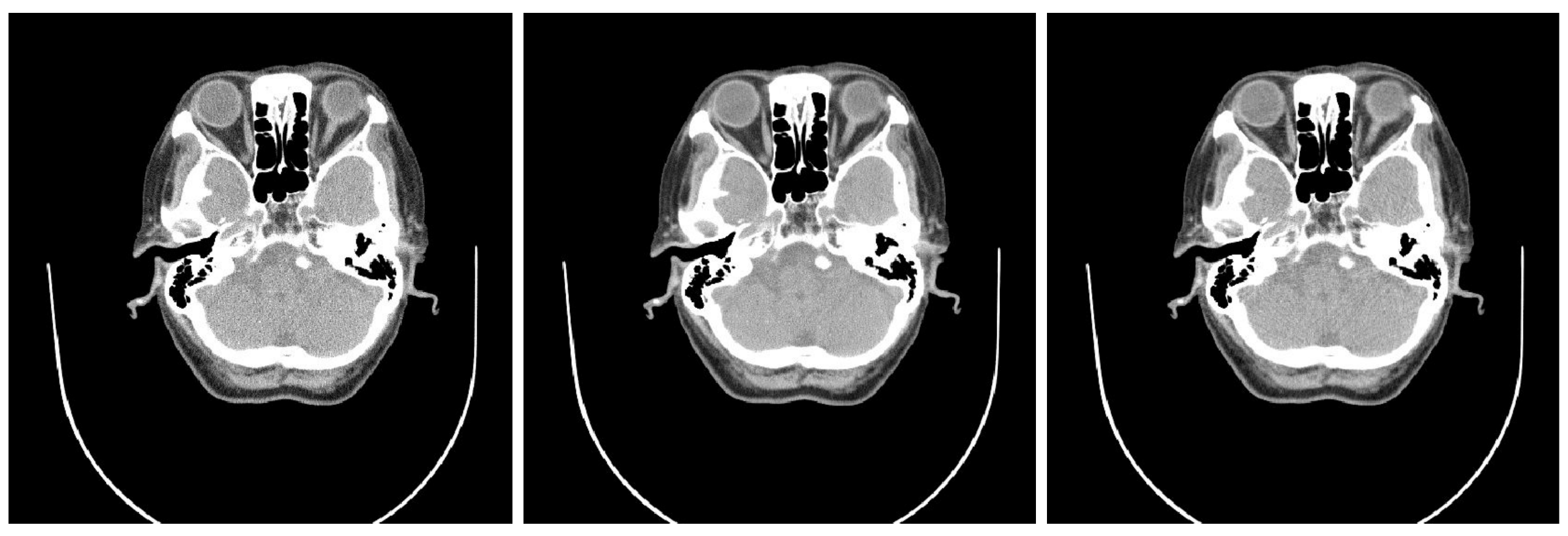}
 \caption{FBP reconstruction from filtered sonogram with low tube current (from left to right): results from the sinograms with 120mA and 400mA, and the filtered sinogram by our method, respectively.}\label{fig:phantom2}
\end{figure}

Finally, we examine our filtering method on real sinogram data by scanning a head phantom in the protocols of both high tube current (400mA) and low one (120mA). In Fig.\ref{fig:phantom2}, fan-beam FBP reconstructed images are shown from original sinograms with different tube currents and the filtered sinogram by our method, respectively. One can observe that, even if in the protocol of a low tube current the FBP reconstructed image from the filtered sinogram by our method can also obtain a better noise-resolution tradeoff with less noise and higher resolution, which is very close to the reconstructed image from the sinogram with high tube current.

In a word, based on above experiments and data analyses, it is verified that the sinogram filtering is an effective method in low-dose CT imaging, and our proposed filtering performs better than related methods in removing noise and artifacts and preserving important details of sinogram data. More important, as a fast data filter our method can be used in the realtime imaging process.

\section{Conclusions}

A fast sinogram filtering based on the LLMMSE estimation is proposed for realtime low-dose X-ray CT imaging in the framework of classic filtered backprojection reconstruction. Developing the nonstationary Gaussian noise model of low-dose sinogram data, and the moving sum technique to speed the algorithm, the proposed fast filtering performs better in noise-resolution tradeoff for reconstructed images, which is verified in the experiments on simulated and real data compared with related filtering methods.

For future research, we will further try to optimize the algorithm in the process of adaptive filtering according to the statistics of noise.

\vskip3pt
\ack{The research has been supported in part by the National Natural Science Foundation of China (61272239, 61070094, 61020106001), the NSFC Joint Fund with Guangdong (U1201258), the Science and Technology Development Project of Shandong Province of China (2014GGX101024), and the Fundamental Research Funds of Shandong University (2014JC012).}

\vskip5pt

\noindent Fengling Wang (\textit{College of Arts Management, Shandong University of Arts, Jinan 250300, China})\\
\noindent Bowen Lin and Shujun Fu (\textit{School of Mathematics, Shandong University, Jinan 250100, China})\\
\noindent Shiling Xie (\textit{Shandong Shtars Biological Industry Co. Ltd, Jinan 250100, China})\\
\noindent Zhigang Zhao (\textit{Shandong Computer Science Center (National Supercomputer Center in Jinan), Qilu University of Technology (Shandong Academy of Sciences), Jinan 250101, China})\\
\noindent Yuliang Li (\textit{Department of Intervention Medicine, The Second Hospital of Shandong University, Jinan 250033, China})

\vskip3pt

\noindent E-mail: Shujunfu@163.com

\end{document}